\documentclass[10pt,twocolumn,letterpaper]{article}

\usepackage{iccv}              % To produce the CAMERA-READY version
\definecolor{iccvblue}{rgb}{0.21,0.49,0.74}
\usepackage[pagebackref,breaklinks,colorlinks,allcolors=iccvblue]{hyperref}
\usepackage[accsupp]{axessibility}  % Improves PDF readability for those with disabilities.

\def\paperID{*****} % *** Enter the Paper ID here
\def\confName{ICCV}
\def\confYear{2025}

\title{The 9th AI City Challenge}

\author{
Zheng Tang$^{1}$ \quad
Shuo Wang$^{1}$ \quad
David C. Anastasiu$^{2}$ \quad
Ming-Ching Chang$^{3}$ \\
Anuj Sharma$^{4}$ \quad
Quan Kong$^{5}$ \quad
Norimasa Kobori$^{5}$ \quad
Munkhjargal Gochoo$^{6,11}$ \\
Ganzorig Batnasan$^{6}$ \quad
Munkh-Erdene Otgonbold$^{6,11}$ \quad
Fady Alnajjar$^6$ \quad
Jun-Wei Hsieh$^7$ \\
Tomasz Kornuta$^1$ \quad
Xiaolong Li$^1$ \quad
Yilin Zhao$^1$ \quad
Han Zhang$^1$ \\
Subhashree Radhakrishnan$^1$ \quad
Arihant Jain$^1$ \quad
Ratnesh Kumar$^1$ \quad
Vidya N. Murali$^1$ \\
Yuxing Wang$^1$ \quad
Sameer Satish Pusegaonkar$^1$ \quad
Yizhou Wang$^1$ \quad
Sujit Biswas$^1$ \\
Xunlei Wu$^1$ \quad
Zhedong Zheng$^8$ \quad
Pranamesh Chakraborty$^9$ \quad
Rama Chellappa$^{10}$ \\
\\
$^1$NVIDIA Corporation, CA, USA \quad
$^2$Santa Clara University, CA, USA \\
$^3$University at Albany, SUNY, NY, USA \quad
$^4$Iowa State University, IA, USA \\
$^5$Woven by Toyota, Japan \quad
$^6$United Arab Emirates University, UAE \\
$^7$National Yang-Ming Chiao-Tung University, Taiwan \\
$^8$University of Macau, Macau \quad
$^9$Indian Institute of Technology Kanpur, India \\
$^{10}$Johns Hopkins University, MD, USA \quad
$^{11}$Emirates Center for Mobility Research, UAE
}

\graphicspath{{./AICity-Org-CVPRW2024}}

\begin{document}
\maketitle
\begin{abstract}
The ninth AI City Challenge continues to advance real-world applications of computer vision and AI in transportation, industrial automation, and public safety. The 2025 edition featured four tracks and saw a 17\% increase in participation, with 245 teams from 15 countries registered on the evaluation server. Public release of challenge datasets led to over 30,000 downloads to date. Track 1 focused on multi-class 3D multi-camera tracking, involving people, humanoids, autonomous mobile robots, and forklifts, using detailed calibration and 3D bounding box annotations. Track 2 tackled video question answering in traffic safety, with multi-camera incident understanding enriched by 3D gaze labels. Track 3 addressed fine-grained spatial reasoning in dynamic warehouse environments, requiring AI systems to interpret RGB-D inputs and answer spatial questions that combine perception, geometry, and language. Both Track 1 and Track 3 datasets were generated in NVIDIA Omniverse. Track 4 emphasized efficient road object detection from fisheye cameras, supporting lightweight, real-time deployment on edge devices. The evaluation framework enforced submission limits and used a partially held-out test set to ensure fair benchmarking. Final rankings were revealed after the competition concluded, fostering reproducibility and mitigating overfitting. Several teams achieved top-tier results, setting new benchmarks in multiple tasks.
\end{abstract}    
\section{Introduction}

The AI City Challenge, in its 9th edition and hosted at ICCV 2025, continued to drive advancements in computer
vision and artificial intelligence (AI) for real-world, high-impact domains. Building on the momentum of previous
years, the 2025 Challenge emphasized scalable and actionable AI solutions across smart transportation, industrial
automation, and public safety.

The 2025 Challenge expanded both scope and ambition, introducing significant upgrades in data scale, modality, and
task complexity. Participants were invited to develop algorithms that analyzed multi-modal sensor data in traffic scenes
and high-fidelity synthetic videos in indoor environments under real-time or near-real-time constraints.

The 2025 edition featured four tracks that reflected critical frontiers in applied AI:
\begin{itemize}
    \item \textbf{Multi-Camera 3D Perception:} This task advanced multi-camera tracking into the 3D domain using a large-scale synthetic dataset generated via NVIDIA Omniverse. Participants tracked diverse object types---including people, service robots, and forklifts---across complex indoor layouts. 3D bounding boxes and camera calibration data were provided. Higher Order Tracking Accuracy (HOTA)~\cite{luiten2020IJCV} was used as the evaluation metric, with a 10\% bonus for online methods that used only past frames.

    \item \textbf{Traffic Safety Description and Analysis:} Participants were challenged to perform detailed video captioning and video question answering (VQA) on staged traffic scenarios involving pedestrian accidents. Using multi-view videos with 3D gaze, bounding boxes, and fine-grained segment annotations, teams generated structured descriptions and answered reasoning questions. Evaluation was based on caption fidelity (BLEU, METEOR, ROUGE-L, CIDEr) and VQA accuracy, with final rankings determined by the mean of both scores.

    \item \textbf{Warehouse Spatial Intelligence:} This track introduced a novel benchmark for fine-grained spatial reasoning in dynamic warehouse environments. Using scene-level RGB-D images and region-specific question-answer pairs, participants developed AI systems that measured object distances, assessed global layouts such as object relationships and distribution, and answered natural language spatial queries. Submissions were evaluated using a weighted success rate across tasks such as counting, measuring, and spatial relation inference.

    \item \textbf{Road Object Detection in Fish-Eye Cameras:} Fish-eye cameras offered panoramic coverage critical for traffic monitoring, but introduced unique challenges due to image distortion. This task, in its second year using the FishEye8K dataset~\cite{Gochoo_2023_CVPR}, assessed real-time object detection frameworks optimized for edge deployment on NVIDIA Jetson devices. Submissions were required to achieve at least 10 FPS on the Jetson AGX Orin 64GB device and were ranked based on the harmonic mean of F1-score and normalized frame rate. Winners were determined through evaluation on an in-house dataset to ensure fairness.
\end{itemize}

The 9th edition of the AI City Challenge generated strong interest and participation, maintaining the momentum established in previous years. Notably, in contrast to earlier editions where all datasets were password-protected, in 2025 several datasets---such as those for Track 1 and Track 3---were made publicly available via Hugging Face. This shift significantly broadened access, resulting in over 30,000 downloads before the challenge submission deadline, a substantial increase compared to the 726 participation requests received in 2024, thereby enhancing the overall impact of the challenge.

Since the announcement of the challenge tracks in late April, the number of teams registered on the evaluation server grew to 245, representing a 17\% increase from the 209 teams in 2024. These teams spanned 15 countries and regions worldwide. Participation across the four challenge tracks was distributed as follows: 11 teams in Track 1, 21 in Track 2, 9 in Track 3, and 91 in Track 4.

This paper presents a comprehensive summary of the preparation and results of the 2025 AI City Challenge. The following sections describe the challenge setup (\S2), dataset preparation (\S3), evaluation methodology (\S4), analysis of participant submissions (\S5), and conclude with a discussion of the findings and future research directions (\S6).

\section{Challenge Setup}
\label{sec:challenge:setup}

The 9th AI City Challenge followed a structured timeline, with the training and validation datasets released on April 30, 2025. This was followed by the release of the test datasets and the activation of the evaluation server on May 30, 2025. Participants were required to submit their challenge results by June 30, 2025, at 6:00 PM Pacific Time. For teams intending to publish accompanying papers, the workshop paper submission deadline was set for July 6, 2025, at 11:59 PM (Anywhere on Earth).

To promote transparency and reproducibility, teams competing for top rankings were required to publicly release their code. This policy ensured that leaderboard results could be independently verified and contributed to the broader research community.

\textbf{Track 1: Multi-Camera 3D Perception.} Participants were tasked with tracking people and mobile objects across more than 500 synthetic camera views generated via NVIDIA Omniverse. The 2025 dataset significantly expanded scene diversity and annotation fidelity, providing 2D and 3D bounding boxes, depth maps, and detailed calibration metadata. Evaluation used the 3D Higher Order Tracking Accuracy (HOTA) metric, and submissions employing online tracking—using only past-frame data—were awarded a 10\% bonus.

\textbf{Track 2: Traffic Safety Description and Analysis.} This track used the Woven Traffic Safety (WTS) dataset~\cite{kong2024wtspedestriancentrictrafficvideo}, which included over 1,200 staged and 4,800 real-world pedestrian-related traffic videos. Each scenario featured multi-view recordings and detailed captions describing pedestrian and vehicle behavior. The 2025 update introduced 2D/3D gaze and head pose annotations using Tobii Pro Glasses, enabling deeper insights into attention and risk. In addition, a Traffic VQA with 180 types of questions (e.g., direction, position, action, attributes) was included along with captioning results as holistic structured information. Participants generated captions and answered visual questions, evaluated using both traditional NLP metrics and a Large Language Model (LLM)-based semantic scorer.

\textbf{Track 3: Warehouse Spatial Intelligence.} Participants answered spatial reasoning questions in warehouse environments using synthetic RGB-D imagery. The dataset included around 500,000 VQA pairs generated via NVIDIA Omniverse, covering four categories: spatial relationships (e.g., left/right), object counting, distance estimation, and multiple-choice identification. Questions were grounded in 3D scene geometry, with annotations including object masks and semantic metadata. The primary evaluation metric was accuracy against normalized ground-truth answers.

\textbf{Track 4: Road Object Detection in Fish-Eye Cameras.} The second-year track used FishEye8K~\cite{Gochoo_2023_CVPR}, FishEye1K\_eval, and an in-house test set to evaluate road object detection under strong lens distortion, varied scales, and lighting, with annotations in VOC, COCO, and YOLO formats. The 2025 edition added an optional Jetson track, where teams submitted TensorRT-optimized Docker containers for real-time benchmarking on the NVIDIA Jetson AGX Orin (64GB) using a standardized script with sequential inference only. Performance was measured in FPS (excluding loading and I/O) and accuracy, with the main metric being the harmonic mean of F1-score and normalized FPS. Winning solutions were required to achieve at least 10 FPS on Jetson.

In summary, the 9th AI City Challenge was carefully structured to balance fairness, transparency, and innovation, ensuring that participating teams could demonstrate the robustness and efficiency of their solutions across diverse, real-world-inspired scenarios.

\section{Datasets}
\label{sec:dataset}

The datasets for the four challenge tracks of the 9th AI City Challenge are described as follows.

%%%%%%%%%%%%%%%%%%%%%%%%%%%%%%%%%%%%%%%%%%%%%%

\subsection{Physical AI Smart Spaces Dataset}

\begin{figure}[t]
\centering
\includegraphics[width=0.47\textwidth]{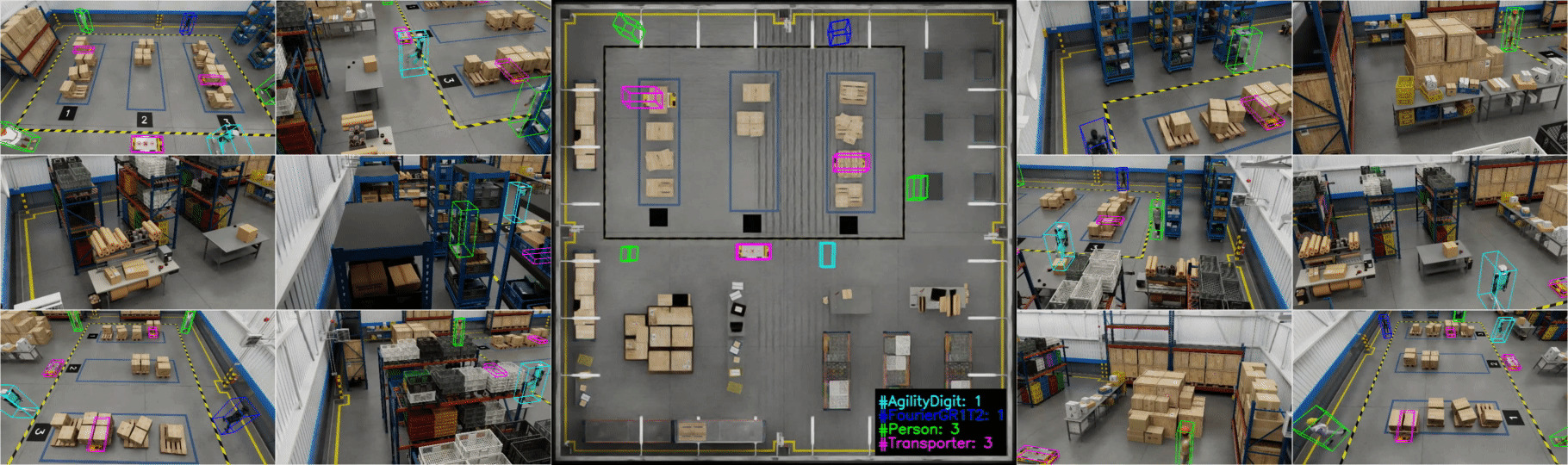}
\caption{Example multi-view annotations from the Physical AI Smart Spaces dataset. The dataset features synchronized multi-camera scenes in a synthetic warehouse, with 2D/3D bounding boxes for various object categories including humans and mobile robots. This high-fidelity dataset was generated using Omniverse Replicator extensions (IRA and IRO).}
\label{fig:physical_ai_smart_spaces}
\end{figure}

The Physical AI Smart Spaces Dataset~\cite{PhysicalAI_SmartSpaces2025} is a large-scale synthetic dataset generated via NVIDIA Omniverse for multi-camera 2D/3D person tracking. It features over 250 hours of video across nearly 1,500 cameras deployed in diverse indoor environments such as warehouses, hospitals, and retail spaces. The dataset includes synchronized videos, 2D/3D bounding boxes, camera calibration files, and depth maps. For 2025, annotations were provided in extended JSON formats, with ground truth expressed in global and per-camera coordinates. The dataset enabled evaluation on 3D HOTA metrics and supported research in large-scale spatial tracking with realistic occlusions and camera overlap.

The dataset was constructed using both the \texttt{isaacsim.replicator.agent} (IRA)~\cite{isaacsim_replicator_agent_2025} and \texttt{isaacsim.replicator.object} (IRO)~\cite{isaacsim_replicator_object_2025} extensions. IRA was used to configure mobile agents with mounted sensors that navigated the environment along realistic trajectories, enabling coverage across blind spots, occlusions, and dynamic viewpoints. IRO facilitated procedural scene composition, including randomized placement of shelving, pallets, and people, while maintaining semantic consistency. The combined use of these extensions enabled the generation of diverse, high-fidelity synthetic scenes suitable for multi-camera tracking, occlusion handling, and spatiotemporal reasoning.

%%%%%%%%%%%%%%%%%%%%%%%%%%%%%%%%%%%%%%%%%%%%%%
\subsection{The Woven Traffic Safety Dataset}

\begin{figure}[t]
\centering
\includegraphics[width=0.47\textwidth]{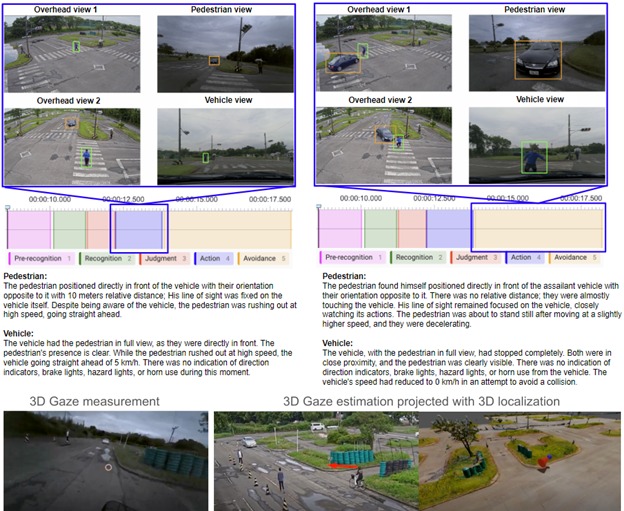}
    \caption{Example annotations from the WTS dataset. Top: synchronized multi-view captures of a pedestrian-involved traffic accident, labeled by semantic phases (Pre-recognition, Recognition, Judgment, Action, and Avoidance). Bottom: visualization of a pedestrian’s visual attention—2D gaze captured via Tobii Pro Glasses (left) and its corresponding 3D projection within the mapped scene (right).}
\label{fig:wts}
\end{figure}

The WTS dataset~\cite{kong2024wtspedestriancentrictrafficvideo} supports fine-grained reasoning over pedestrian and vehicle interactions using synchronized video streams from overhead, vehicle-mounted, and ego-centric perspectives. It includes over 1,200 staged traffic events and 4,800 real-world videos filtered from BDD100K. Each scenario was temporally aligned and richly annotated with multi-sentence captions structured around five semantic phases: Pre-recognition, Recognition, Judgment, Action, and Avoidance.

The 2025 edition introduced several major enhancements. Most notably, it added 2D and 3D gaze annotations and head pose tracking for key pedestrian actors, collected using Tobii Pro Glasses and aligned to a pre-mapped 3D model of each scene. These additions enabled the study of pedestrian attention, risk awareness, and situational understanding in dynamic environments. In addition to captioning, each video was accompanied by VQA tasks. Responses were evaluated using both standard NLP metrics and a LLM-based scoring framework to assess semantic alignment with human annotations.

%%%%%%%%%%%%%%%%%%%%%%%%%%%%%%%%%%%%%%%%%%%%%%
\subsection{Physical AI Spatial Intelligence Warehouse Dataset}
\begin{figure}[t]
\centering
\includegraphics[width=0.48\textwidth]{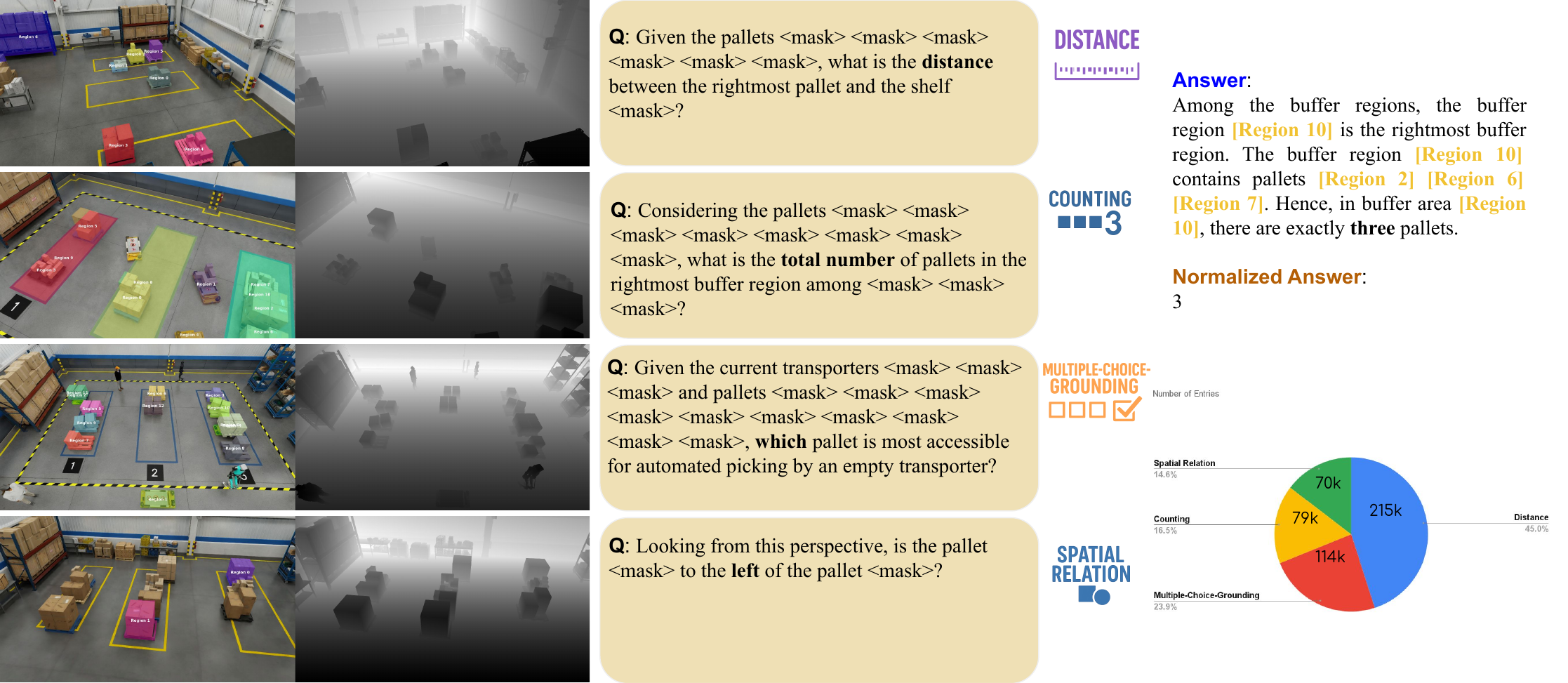}
\caption{Example images from the Physical AI Spatial Intelligence Warehouse dataset. Tasks included \textbf{distance}, \textbf{counting}, \textbf{multiple-choice grounding}, and \textbf{spatial relation reasoning}. Each example shows the RGB frame, depth map, annotated regions, the corresponding question, and sample answers. The distribution of question types demonstrated the diversity of reasoning skills required across tasks.}
\label{fig:warehouse_spatial_intelligence}
\end{figure}

This dataset~\cite{PhysicalAI_SpatialIntelWarehouse2025} focuses on spatial reasoning and vision-language understanding in synthetic warehouse environments. It contains around 500,000 VQA samples generated via an automated pipeline leveraging Omniverse simulations and auto-labeling engines, covering four key task types: spatial relationship queries (e.g., left/right/above), object counting, distance measurement, and multiple-choice identification. Each sample includs RGB-D imagery, object masks, natural-language conversation pairs with questions, and both free-form and normalized textual answers aligned to a structured ontology. Ground-truth annotations adopt the common LLaVA~\cite{liu2023visual} format to support multimodal learning and spatial reasoning.

The dataset was generated using both IRA and IRO extensions~\cite{isaacsim_replicator_agent_2025, isaacsim_replicator_object_2025} in Omniverse Isaac Sim. IRO enabled YAML-driven configuration of object layouts, including 3D bounding boxes, 2D bounding boxes, 2D masks, and semantic class tagging across warehouse elements. IRA provided dynamic and fixed camera agent controls, allowing views from robot-mounted or stationary perspectives. This combination produced richly varied camera placements and systematic scene randomization. Based on the ground-truth geometry, normalized answers in human-readable units were computed. An optional stage of answer diversification was applied via LLM-based rephrasing. The result was a scalable, reproducible benchmark for training and evaluating spatial intelligence models in complex indoor spaces.

%%%%%%%%%%%%%%%%%%%%%%%%%%%%%%%%%%%%%%%%%%%%%%
\subsection{The {\bf \textit{FishEye8K}} and {\bf \textit{FishEye1Keval}} Datasets}

\begin{figure}[t]
\centering
\includegraphics[width=0.47\textwidth]{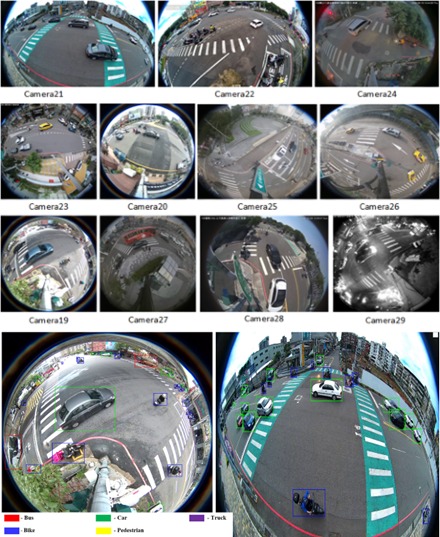}
\caption{Example images from the FishEye8K dataset. Top: one representative frame from each of the 18 surveillance-grade fisheye cameras used to collect the dataset, covering both day and night scenes in urban traffic intersections. Bottom: sample object detection results highlighting bounding boxes across five object categories (Bike, Car, Pedestrian, Truck, and Bus) captured from challenging wide-angle views.}
\label{fig:fisheye8K}
\end{figure}

The FishEye8K dataset~\cite{Gochoo_2023_CVPR} was utilized for training and validation sets, while FishEye1K\_eval, the evaluation set from the 2024 challenge, was used as part of the test set for the tentative leaderboard. Although the labels of FishEye1K\_eval were never released, using it to determine challenge winners for a second year was deemed unfair. Thus, the organizers curated a separate in-house dataset to decide the challenge winners.

The FishEye8K dataset, originally introduced in~\cite{Gochoo_2023_CVPR}, consists of 8,000 annotated images captured from 18 wide-angle surveillance cameras installed in Hsinchu City, Taiwan. The dataset is split into 5,288 training and 2,712 validation images at resolutions of 1080$\times$1080 and 1280$\times$1280, respectively, with a total of 157K bounding boxes across five object categories: Bus, Bike, Car, Pedestrian, and Truck. The labels are available in multiple formats, including PASCAL VOC, COCO, and YOLO. The FishEye1K\_eval test set included 1,000 images sourced from 11 camera streams. Both the FishEye1K\_eval and the in-house test set used for final evaluation were extracted from streams recorded in the same city as the FishEye8K dataset.
\section{Evaluation Protocols}
\label{sec:evaluation}

The 9th AI City Challenge featured four tracks spanning multi-camera 3D tracking, pedestrian traffic safety analysis, spatial reasoning in warehouses, and road object detection in fisheye cameras. The datasets, task objectives, submission formats, and evaluation metrics are summarized below.

\subsection{Track 1 Evaluation}

Track 1 challenged teams to detect and consistently track 3D objects across multiple synchronized cameras in complex indoor environments~\cite{PhysicalAI_SmartSpaces2025}. The dataset was generated with synthetic animated scenes using both the IRA and IRO extensions in NVIDIA Omniverse. The 2025 edition expanded to 42 hours of video across 504 cameras in 19 diverse indoor layouts (warehouses, hospitals, retail stores, offices). Over 360 unique object instances (people, robots, forklifts) were labeled.

Each scene provided RGB video, 3D/2D annotations, camera calibrations, and optional depth maps. Participants were required to track object identities across views and time, submitting one detection per line in the following format:

\texttt{<scene\_id> <class\_id> <object\_id> <frame\_id> <x> <y> <z> <width> <length> <height> <yaw>}

The evaluation metric was 3D HOTA~\cite{luiten2020IJCV}, which jointly measured detection, association, and localization quality. For each scene, HOTA scores were computed per class, averaged, and then aggregated across all scenes with weights proportional to the number of objects. Ground-truth and predicted objects were matched using the 3D IoU of bounding boxes, a stricter criterion than the previous year’s 2D center-distance approach. For award consideration, online methods (relying only on past frames) received a 10\% bonus.

\subsection{Track 2 Evaluation}

Track 2 evaluated temporal understanding in pedestrian traffic accidents using the WTS dataset~\cite{kong2024wtspedestriancentrictrafficvideo}, which included 810 multi-view videos of 155 traffic scenarios staged with stunt actors and real vehicles. Each event was segmented into five phases: Pre-recognition, Recognition, Judgment, Action, and Avoidance. Detailed captions were provided per phase for both pedestrians and vehicles, along with gaze, head pose, and VQA annotations.

The task consisted of two subtasks: (1) generating pedestrian and vehicle captions for each segment, and (2) answering multiple-choice questions on spatial/behavioral details. Evaluation used BLEU-4, METEOR, ROUGE-L, and CIDEr for captions, and accuracy for VQA. Final rankings averaged both subtask scores.

\subsection{Track 3 Evaluation}

Track 3 targeted spatial reasoning in industrial warehouse environments using the Physical AI Spatial Intelligence Warehouse dataset~\cite{PhysicalAI_SpatialIntelWarehouse2025}. Generated using both IRA and IRO extensions, the dataset included 3D object layouts and language-based spatial reasoning questions.

Participants submitted answers to questions such as object grounding, distances, spatial relations, and counts. Success was measured via accuracy or Acc@10 depending on question type, with a weighted average used for leaderboard ranking. A strict normalization scheme ensured consistency in answer evaluation.

\subsection{Track 4 Evaluation}

Track 4 focused on road object detection in wide-angle fisheye images, with a two-stage evaluation process. Initially, teams submitted predictions in COCO JSON format on the FishEye1K\_eval dataset for a tentative leaderboard ranked by F1-score. For the final evaluation, teams competing for prizes were required to provide a Docker container optimized for real-time inference on the Jetson AGX Orin (64GB). Submissions failing to achieve at least 10 FPS on the evaluation hardware were disqualified. The final ranking was determined by the harmonic mean of the F1-score and normalized FPS on the in-house test set, prioritizing both lightweight and accurate solutions,

\begin{equation}
\text{Metric} = \frac{2 \cdot F1 \cdot \text{NormFPS}}{F1 + \text{NormFPS}},
\end{equation}

\begin{equation}
\text{NormFPS} = \frac{\min(\text{FPS}, \text{MaxFPS})}{\text{MaxFPS}},
\end{equation}

where MaxFPS = 25.

\subsection{2025 Evaluation System}

The 2025 AI City Challenge used a centralized evaluation portal. Participants created an evaluation account with an email and password and verified their email to activate the account. After verification, submission access was granted by an administrator. Once approved, teams could submit to individual tracks through the ``Add'' button under the Submissions tab. Each submission could be directed to either the \textbf{Public} leaderboard—eligible for awards and publication—or the \textbf{General} leaderboard.

To qualify for awards via the Public leaderboard, teams were required to:
\begin{itemize}
\item Use no external private data in training or evaluation.
\item Release reproducible code, trained models, and any created labels before the deadline.
\end{itemize}

Submission limits were set as follows:
\begin{itemize}
\item Up to 5 submissions per track per day (Pacific Time).
\item Maximum 30 submissions for Track 1, 20 for Tracks 2 and 3, and 50 for Track 4.
\end{itemize}

Scores on the leaderboard during the active challenge period reflected performance on a 50\% subset of the test set. Additionally, only the top 3 team scores (and the user’s own team’s current rank) were visible. After the challenge deadline, the results were automatically updated to reflect evaluation on the full test set. Final leaderboards displayed all team names and scores.
\section{Challenge Results}
\label{sec:results}

Tables~\ref{table:1}–\ref{table:4} provide a summary of the leaderboard results for Tracks 1 through 4, respectively.

%%%%%%%%%%%%%%%%%%%%%%%%%%%%%%%%%%%%%%%%%%%
\subsection{Summary for the Track 1 Challenge}

\begin{table}[t]
\caption{Summary of the Track 1 leader board.}
\label{table:1}
\centering
\footnotesize
\begin{tabular}{|c|c|c|c|c|}
\hline
Rank & Team ID & Team & Score & Online \\
\hline\hline
1 & 65 & ZV~\cite{AICity25Paper21} & {\bf 69.9118} & No \\
\hline
2 & 15 & SKKU-AutoLab~\cite{AICity25Paper12} & 63.1396 & No \\
\hline
3 & 133 & TeamQDT~\cite{AICity25Paper26} & 28.7515 & Yes \\
\hline
4 & 116 & UTE AI Lab~\cite{AICity25Paper23} & 25.3983 & Yes \\
\hline
\end{tabular}
\vspace{-0.4cm}
\end{table}

The 2025 Track 1 Challenge focused on 3D Multi-Target Multi-Camera (MTMC) tracking in large-scale synthetic indoor environments. Participants were tasked with estimating 3D bounding boxes and maintaining identity consistency across disjoint views, with emphasis on handling occlusion and class imbalance. The top submissions reflected two dominant strategies: offline geometry-first pipelines and online tracking-first systems.

The winning team, ZV~\cite{AICity25Paper21}, adopted a fully offline, geometry-centric approach by fusing depth maps into a unified 3D point cloud and using a transformer-based V-DETR for detection. Their Embedding Similarity Classifier (ESC) addressed rare class imbalance by leveraging 3D shape embeddings, and a hybrid tracking framework combined short-term online tracking with global offline association.

SKKU-AutoLab~\cite{AICity25Paper12} introduced DepthTrack, a modular offline pipeline connecting BEV-space tracklets with clustered point clouds via Tracklet-Cluster Mapping (TCM). Their system integrated ViT-based ReID, pose-based yaw estimation, and low-light enhancement to improve robustness across visual conditions.

TeamQDT~\cite{AICity25Paper26} proposed an efficient online framework extending 2D MTMC tracking into 3D via late-stage depth fusion. Using MOT ID consistency, agglomerative clustering, and footpoint projection, their system achieved strong performance with low overhead and real-time applicability.

UTE AI Lab~\cite{AICity25Paper23} also employed an online approach with a focus on 3D box refinement. Their View-Aware Geometric Center Refinement (VGCR) improved localization accuracy, and novel affinity metrics—such as trajectory-level Fréchet distance and 3D IoU—enhanced identity association under occlusion.

All teams leveraged depth and calibration data, but with differing granularity: ZV and SKKU-AutoLab performed early fusion and sparse 3D reasoning, while TeamQDT and UTE AI Lab used lighter frame-wise projections. ViT-based ReID, pose-guided yaw estimation, and modular clustering were widely adopted. While offline methods led in accuracy, online pipelines showed promising trade-offs for real-time deployment.

%%%%%%%%%%%%%%%%%%%%%%%%%%%%%%%%%%%%%%%%%%%
\subsection{Summary for the Track 2 Challenge}

\begin{table}[t]
  \caption{Summary of the Track 2 leader board.}
  \label{table:2}
  \centering
  \footnotesize
  \begin{tabular}{|c|c|c|c|}
    \hline
    Rank & Team ID & Team & Score \\
    \hline\hline
    1 & 145 & CHTTLIOT~\cite{AICity25Paper4} & {\bf 60.0393} \\
    \hline
    2 & 1 & SCU\_Anastasiu~\cite{AICity25Paper34} & 59.1184 \\
    \hline
    3 & 52 & Metropolis\_Video\_Intelligence~\cite{AICity25Paper14} & 58.8483 \\
    \hline
    4 & 137 & ARV~\cite{AICity25Paper13} & 57.9138 \\
    \hline
    5 & 121 & Rutgers ECE MM~\cite{AICity25Paper27} & 57.4658 \\
    \hline
    6 & 68 & VNPT\_AI~\cite{AICity25Paper22} & 57.1133 \\
    \hline
    7 & 60 & BAO\_team~\cite{AICity25Paper42} & 55.6550 \\
    \hline
    10 & 49 & MIZSU~\cite{AICity25Paper16} & 45.7572 \\
    \hline
  \end{tabular}
\end{table}

The 2025 Track 2 Challenge emphasized fine-grained video captioning and safety-focused VQA from multi-camera traffic footage. Top teams leveraged large Vision-Language Models (VLMs) with domain-specific enhancements such as spatiotemporal prompt engineering, multi-agent specialization, and lightweight fine-tuning techniques.

The winning team, CHTTLIOT~\cite{AICity25Paper4}, developed TrafficInternVL, which used keyframe-based global–local views, role-aware prompts, and joint QA–VQA training.

SCU\_Anastasiu~\cite{AICity25Paper34} achieved second place with a modular multi-agent framework that routed tasks to specialized agents based on validation performance.

Metropolis\_Video\_Intelligence~\cite{AICity25Paper14} introduced TrafficVILA, a high-resolution system incorporating dynamic tiling, SAM2-based prompting, and fact-checking to improve scene fidelity.

Other top entries included ARV~\cite{AICity25Paper13}, which combined spatially guided prompting with caption refinement; Rutgers ECE MM~\cite{AICity25Paper27}, which used phase-aware view selection and key moment identification; and VNPT\_AI~\cite{AICity25Paper22}, which injected expert cues such as gaze and attention for enhanced VQA. BAO team~\cite{AICity25Paper42} focused on lightweight caption decomposition and view filtering, while MIZSU~\cite{AICity25Paper16} adopted a dual-model setup that separated captioning and VQA for improved specialization.

Overall, most teams fine-tuned open-source VLMs like Qwen2.5-VL and InternVL, with shared emphasis on prompt engineering, frame selection, and hierarchical captioning. Offline methods dominated in accuracy, but several entries also demonstrated promising directions for efficient, real-world VLM deployment in traffic safety systems.

%%%%%%%%%%%%%%%%%%%%%%%%%%%%%%%%%%%%%%%%%%%
\subsection{Summary for the Track 3 Challenge}

\begin{table}[t]
\caption{Summary of the Track 3 leader board.}
\label{table:3}
\centering
\footnotesize
\begin{tabular}{|c|c|c|c|}
\hline
Rank & Team ID & Team & Score \\
\hline\hline
1 & 16 & UWIPL\_ETRI~\cite{AICity25Paper3} & {\bf 95.8638} \\
\hline
2 & 57 & HCMUT.VNU~\cite{AICity25Paper20} & 91.9735\\
\hline
4 & 140 & Embia~\cite{AICity25Paper9} & 90.6772\\
\hline
4 & 49 & MIZSU~\cite{AICity25Paper11} & 73.0606\\
\hline
5 & 99 & HCMUS\_HTH~\cite{AICity25Paper41} & 66.8861\\
\hline
\end{tabular}
\vspace{-0.4cm}
\end{table}

The 2025 Track 3 Challenge addressed spatial reasoning in warehouse environments using VLMs or agentic solutions. Participants were tasked with developing multimodal systems capable of understanding geometric relationships—such as proximity, inclusion, and orientation—based on RGB, depth, and textual inputs. Leading approaches emphasized lightweight architectures, region-aware reasoning modules, and explicit depth integration to balance spatial accuracy and runtime efficiency.

The winning team, UWIPL\_ETRI~\cite{AICity25Paper3}, proposed a modular agent framework combining LLMs with tool-augmented spatial APIs. Their method used dedicated modules for perception, region localization, and multi-turn reasoning, enabling precise interpretation of spatial relationships with minimal fine-tuning.

HCMUT.VNU~\cite{AICity25Paper20} introduced an enhanced SpatialRGPT architecture incorporating a Region Feature Enhancer and a dedicated Region Classifier. Their multi-stage training pipeline improved RGB-depth feature fusion and offloaded classification tasks from the LLM, boosting spatial grounding accuracy.

Embia~\cite{AICity25Paper9} developed SmolRGPT, a compact 600M-parameter architecture that retained strong reasoning capabilities while being suitable for deployment in resource-constrained environments. Their design used separate refinement pipelines for RGB and depth features, coupled with a pixel-shuffling-based visual projector to preserve dense spatial cues.

MIZSU~\cite{AICity25Paper11} presented a baseline built on a pretrained GPT model with spatial prompting and region-focused input formatting. Their method emphasized handcrafted preprocessing and prompt tuning to guide spatial reasoning, achieving solid results without the use of large-scale visual encoders.

Finally, HCMUS\_HTH~\cite{AICity25Paper41} introduced TinyGiantVLM, a dual-branch architecture with RGB-depth fusion and a Mixture-of-Experts strategy tailored for low-latency spatial inference.

Collectively, these submissions showcased how spatially grounded reasoning could be achieved through modular tools, depth-aware representations, and parameter-efficient designs. The challenge highlighted emerging trade-offs between scalability, accuracy, and deployability in the evolving landscape of vision-language intelligence.

%%%%%%%%%%%%%%%%%%%%%%%%%%%%%%%%%%%%%%%%%%%
\subsection{Summary for the Track 4 Challenge}

\begin{table}[t]
\caption{Summary of the Track 4 leader board.}
\label{table:4}
\centering
\footnotesize
\begin{tabular}{|c|c|c|c|}
\hline
Rank & Team ID & Team                                 & Score  \\ \hline
1    & 33      & UIT-OpenCubee~\cite{AICity25Paper39} & 0.6493 \\ \hline
2    & 43      & UT\_T1~\cite{AICity25Paper5}         & 0.6413 \\ \hline
4    & 45      & Zacian~\cite{AICity25Paper35}        & 0.6405 \\ \hline
5    & 15      & SKKU-AutoLab~\cite{AICity25Paper17}  & 0.6397 \\ \hline
8    & 67      & Smart Lab~\cite{AICity25Paper7}      & 0.6366 \\ \hline
10   & 5       & SmartVision~\cite{AICity25Paper8}    & 0.6342 \\ \hline
12   & 22      & VGU-VLinsight~\cite{AICity25Paper25} & 0.6318 \\ \hline
13   & 86      & Tyche~\cite{AICity25Paper32}         & 0.6302 \\ \hline
14   & 53      & xiilab~\cite{AICity25Paper33}        & 0.6268 \\ \hline
\end{tabular}
\vspace{-0.4cm}
\end{table}

Track 4 of the 2025 AI City Challenge focused on object detection from single fisheye cameras under real-time constraints, requiring high accuracy despite severe radial distortion and the limitations of edge-device inference. Most of the top-performing methods on the tentative leaderboard leveraged data-centric strategies, lightweight model architectures, and distortion-aware augmentations to address the unique challenges posed by fisheye imagery.

UIT-OpenCubee~\cite{AICity25Paper39}, ranked first on the tentative leaderboard, proposed a modular and lightweight pipeline integrating distortion-aware copy-paste augmentation, synthetic data generation, and dual-model inference using YOLOv11 and D-FINE. Their design emphasized data diversity and geometric consistency while preserving real-time speed through Weighted Boxes Fusion.

UT\_T1~\cite{AICity25Paper5}, in second place, utilized a calibration-free ensemble of three YOLO-based models, trained on diverse domain-specific datasets. They applied gamma-corrected preprocessing to improve detection and used Weighted Boxes Fusion (WBF) to address illumination variations and enhance accuracy.

Zacian~\cite{AICity25Paper35} enhanced a framework from the previous year’s team~\cite{AICity24Paper4}, ensembling multiple YOLO variants and CoDETR models, incorporating fisheye-specific augmentations and pseudo-labeling. They also implemented nighttime simulation using CycleGAN to improve robustness in low-light conditions.

SKKU-AutoLab~\cite{AICity25Paper17} focused exclusively on augmentation, introducing FED, BIDA, and D2N techniques to generate synthetic fisheye training data, paired with the DEIM detector optimized for real-time deployment via mixed-precision quantization. Their system achieved over 25 FPS on Jetson AGX Orin 32GB.

Smart Lab~\cite{AICity25Paper8} proposed a unified pipeline combining simple pre-/post-processing techniques and a four-model ensemble (YOLOR, YOLOv12, Salience-DETR, Co-DETR), fine-tuned for nighttime and peripheral distortions using synthetic augmentation and pseudo-labeling.

SmartVision~\cite{AICity25Paper8} prioritized edge-device readiness through C++/TensorRT optimization and a compact single YOLO11m model, distilled from Co-DETR and trained on barrel-transformed VisDrone images with synthetic labels. The system achieved 63.42\% F1 at 25 FPS on Jetson AGX Orin (64GB). 

VGU-VLinsight~\cite{AICity25Paper25} proposed a multi-stage DETR distillation framework, transferring knowledge from Co-DETR to D-FINE via four progressive phases, enhanced with CycleGAN-based augmentation and adaptive sample mining. Their method achieved 63.18\% F1 at 145 FPS on RTX4090.

Tyche~\cite{AICity25Paper32} introduced APAM and RADA modules to prioritize peripheral learning during training and implemented TensorRT optimizations for deployment without sacrificing speed.

Lastly, Xiilab~\cite{AICity25Paper33} developed a fisheye-aware real-time DFINE system, employing distortion-preserving augmentation and active learning-based pseudo-labeling. The proposed method achieved 62.68\% F1 at over 15 FPS on Jetson AGX Orin (32GB).

Collectively, these submissions highlighted a trend toward scalable, distortion-aware, and hardware-efficient pipelines. By focusing on data diversity, peripheral object modeling, and GPU-aware deployment, the top teams advanced the state of the art in real-time fisheye object detection for intelligent transportation applications.
\section{Discussion and Conclusion}
\label{sec:conclusion}

The 9th AI City Challenge marked a significant milestone in advancing applied computer vision across intelligent transportation, industrial automation, and public safety. With expanded datasets, novel benchmarks, and a 17\% increase in participation, the 2025 competition pushed the boundaries of multimodal scene understanding, 3D spatial reasoning, and real-time perception.

Track~1 introduced a large-scale synthetic benchmark for 3D multi-camera tracking across indoor spaces with people, service robots, and forklifts. Offline methods achieved the highest HOTA scores by leveraging dense fusion and global association, while online pipelines demonstrated promising directions for real-time deployment. A key research opportunity remains in learning-based identity association, especially in handling occlusion and dynamic entries/exits in crowded environments. Future iterations may further challenge generalization with increased visual ambiguity, such as similar clothing or object appearances.

Track~2 addressed fine-grained traffic safety analysis with multi-view pedestrian scenarios enriched by 3D gaze and head pose annotations. Top-performing teams adopted VLMs enhanced by spatio-temporal prompting and role-aware representations. However, current evaluation metrics often fall short in assessing the semantic fidelity of long, structured descriptions. This motivates future research on instance-level video language grounding and LLM-guided caption generation tailored to complex traffic contexts. Beyond understanding, reasoning about the causes of traffic scenarios would be an essential next step.

Track~3 featured a novel dataset on warehouse spatial intelligence with around 500K RGB-D question–answer pairs. Leading submissions combined lightweight visual transformers with region-aware modules and explicit geometric encoding. Modular architectures and tool-augmented reasoning emerged as effective strategies for balancing depth integration and spatial understanding. Continued exploration into efficient VLMs and scalable 3D language grounding is essential for real-world robotics and automation use cases.

Track~4 focused on real-time object detection in fisheye cameras, emphasizing deployment on edge devices. Teams employed distortion-aware augmentations, model ensembling, and Jetson-optimized pipelines. The top solutions demonstrated the importance of synthetic data diversity, lightweight architectures, and quantized inference. Yet, peripheral object accuracy and domain adaptation remain challenging, pointing to future efforts in unified training pipelines for wide-angle video analytics.

Across all tracks, the public release of datasets on Hugging Face led to over 30,000 downloads, fostering broader community engagement. The 2025 Challenge highlighted a strong trend toward multimodal fusion, domain-specific model design, and real-time readiness. As AI systems move toward deployment in safety-critical scenarios, continued convergence of perception, language, and reasoning will drive innovation in intelligent cities.

\section{Acknowledgment}

The datasets for the 9th AI City Challenge were developed through extensive data curation efforts enabled by close collaboration between industry and academia. Key contributors included NVIDIA Corporation and Woven by Toyota, Inc., alongside academic partners such as National Yang Ming Chiao Tung University, United Arab Emirates University, and the Emirates Center for Mobility Research (ECMR), supported in part by Grant 12R012. The challenge also benefited from thorough review efforts by researchers at the University at Albany and the University of Macau. Special thanks are extended to Santa Clara University for their critical contributions to the development and continuous enhancement of the evaluation systems.

We thank the broader research community for their participation, which helped establish the AI City Challenge as a leading benchmark in intelligent transportation and spatial AI. The contributions of participants, reviewers, and collaborators were essential to the success of the 2025 edition.

{
    \small
    \bibliographystyle{ieeenat_fullname}
    \bibliography{aicity17,aicity18,aicity19,aicity20,aicity21,aicity22,aicity23,aicity24,aicity25}
}

\end{document}